\documentclass[3p,authoryear,preprint,review, 11pt]{elsarticle}

\usepackage[utf8x]{inputenc}
\usepackage[T1]{fontenc}

\usepackage{subcaption}

\usepackage{amsmath}
\usepackage{amssymb}
\usepackage{graphicx}
\usepackage[colorinlistoftodos]{todonotes}
\usepackage[colorlinks=true, allcolors=blue]{hyperref}
\usepackage{enumitem}
\usepackage{makecell}
\usepackage{algorithm}
\usepackage{algpseudocode}
\usepackage{fixltx2e}

\usepackage{booktabs}
\usepackage{url}

\usepackage{multicol}
\usepackage{float}
\usepackage{lineno}
\usepackage{setspace}

\journal{N/A}
\usepackage{lipsum}
\usepackage{enumitem}

\begin{document}
\begin{frontmatter}
	
	
	
	\title{End-to-end Deep Reinforcement Learning for Stochastic Multi-objective Optimization in C-VRPTW}
	
	
	\author[TUE]{Abdo Abouelrous\fnref{*}}
	\ead{a.g.m.abouelrous@tue.nl}
        \fntext[*]{corresponding author}
    \author[TUE]{Laurens Bliek}
	  \ead{l.bliek@tue.nl}
          \author[TUE]{Yaoxin Wu}
        \ead{y.wu2@tue.nl}
        \author[TUE]{Yingqian Zhang}
	\ead{yqzhang@tue.nl}
	\address[TUE]{Department of Information Systems, Faculty of Industrial Engineering and Innovation Sciences, Technical University Eindhoven, The Netherlands}

\begin{abstract}

In this work, we consider learning-based applications in routing to solve a Vehicle Routing variant characterized by stochasticity and multiple objectives. Such problems are representative of practical settings where decision-makers have to deal with uncertainty in the operational environment as well as multiple conflicting objectives due to different stakeholders. We specifically consider travel time uncertainty. We also consider two objectives, total travel time and route makespan, that jointly target operational efficiency and labor regulations on shift length, although more/different objectives could be incorporated. Learning-based methods offer earnest computational advantages as they can repeatedly solve problems with limited interference from the decision-maker. 
We specifically focus on end-to-end deep learning models that leverage the attention mechanism and multiple solution trajectories.
These models have seen several successful applications in routing problems.
However, since travel times are not a direct input to these models due to the large dimensions of the travel time matrix, accounting for uncertainty is a challenge, especially in the presence of multiple objectives. In turn, we propose a model that simultaneously addresses stochasticity and multi-objectivity and provide a refined training mechanism for this model through scenario clustering to reduce training time. Our results show that our model is capable of constructing a Pareto Front of good quality within acceptable run times compared to three baselines. We also provide two ablation studies to assess our model's suitability in different settings.
\end{abstract}


\end{frontmatter}

\section{Introduction}
Real-life Combinatorial Optimization (CO) problems, such as routing problems, are often characterized by certain challenging features. Examples of such features include stochasticity and the presence of several (conflicting) objectives simultaneously. In applications like the Vehicle Routing Problem (VRP), accounting for these features during decision-making is important. Otherwise,  sub-optimal  decisions may lead to elevated costs.

Furthermore, computational budgets are often limited in practice as solutions need to be generated quickly in accordance with operational requirements \citep{horvitz2013reasoning}. To that end, one seeks optimization techniques that are not only capable of delivering solutions efficiently but also delivering near-optimal solutions in the presence of the aforementioned features.

In the literature, numerous techniques have been proposed to solve CO problems, ranging from hand-crafted heuristics to Machine Learning (ML) models that are independently able to generate solutions - also known as end-to-end methods. The spectrum also covers hybrid methods, which incorporate both heuristics and ML models. ML models are particularly useful in cases where problem parameters follow a known distribution. In such cases, an ML model can be trained on a dataset of problems from this distribution and used to generate solutions for problem instances arising in the future.

The ability of ML models to generate solutions statistically - even when the solution distribution is unknown beforehand - by leveraging data from the operational environment to make decisions in a principled fashion, has been repeatedly stressed in the literature \citep{bengio2021machine,giuffrida2022optimization,mazyavkina2021reinforcement}. This is because ML models have the capacity to learn from the collective expert knowledge. Through learning common solution structures and relationships to problem parameter input, ML model can continuously solve different problems. Although the training resources required are not trivial, ML models can be quickly applied for solving thereafter \citep{zhang2021solving}. This is in stark contrast to hand-crafted heuristics which, generally, do not leverage information from the solution structure \citep{d2020learning} and may be challenging to repeatedly develop and/or replicate for newer instances \cite{accorsi2022guidelines}.

For the large part, the CO problems treated in the literature by ML models were deterministic and had a single objective. Real-life problems, contrarily, are often characterized by uncertainty and multiple (conflicting) objectives. The objective of our study is to extend the framework of ML in optimization to settings that jointly combine the stochastic and multi-objective features. In our estimation, this is an under-studied application with significant industrial relevance. 

Solving a multi-objective problem requires the definition of a Pareto Front \citep{ngatchou2005pareto}, comprising of a set of Pareto optimal solutions by which improving one objective can not be done without worsening other objectives. Solutions would then have to be validated on a sample of scenario realizations to estimate their quality and feasibility. In an optimization context, this poses a challenge that requires novel intervention. This is because multi-objectivity induces the generation of several solutions to estimate the Pareto Front, while stochasticity imposes that each solution be feasible with a good score for a large number of stochastic realizations. The repeated evaluation of multiple solutions, in this setting, is a cumbersome task that requires careful choices in the optimization methodology to maintain computational costs within practical limits.

In response, we present a framework that addresses the given computational issues in this paper to solve Capacitated Vehcile Routing Problem with Time Windows (C-VRPTW). We specifically focus on stochasticity in travel times and consider two objectives that are travel time-based, although our method could account for more different objectives. The major contributions prescribed by this paper are:
\begin{itemize}[noitemsep]
    \item Presenting the first end-to-end (independently constructs solution without interference of a heuristic) deep-learning model that jointly treats stochasticity and multiple objectives for routing.
    \item Providing a retraining mechanism for parameter uncertainty through an active search that considers multiple objectives rather than just one objective.
    \item Presenting a scenario-clustering technique that enhances the computational efficiency of model retraining during active search.
\end{itemize}

The remainder of the paper is organized as follows. Section \ref{previous work} introduces previous research. Section \ref{Problem Des} describes the relevant problem in detail. Section \ref{method} outlines our methodology. Section \ref{Numerical Experiments} presents numerical experiments and results. Lastly, Section \ref{Conclusionss} asserts the conclusions.

\section{Previous Work}
\label{previous work}

We restrict our attention to ML applications in stochastic and/or multi-objective cases for routing. These models generally rely on concepts of Reinforcement Learning (RL) to make decisions. These approaches fall into two major categories, hybrid and end-to-end. Hybrid models integrate a ML model with a heuristic to enhance the heuristic's performance. The way in which the ML model is applied largely depends on the heuristic at hand. A common example is that the RL agent selects a heuristic with some probability from a set of discrete heuristics. In contrast, End-to-end models are able to independently solve a routing problem without interference of a heuristic. When deciding on the ordering of node visits, the decision to visit a node $j$ after node $i$ is based on a probability $p_{ij}$. Such a probability may be determined by features of nodes $i$ and $j$, the status of the current (partial) solution and the estimated corresponding reward. In both categories, the decisions are determined by interaction of the problem input with the ML model's parameters which are optimized during model training to maximize the final reward.

For the multi-objective setting, works like \cite{wu2024multiobjective} and \cite{deng2024multi} explore applications to routing with hybrid methods through genetic algorithms. \cite{yao2017efficient}, on the other hand, focus on hyper-heuristic selection by the RL agent to complete a solution.  In the Stochastic setting, \cite{bayliss2021machine} make use of simulation-augmented optimization for urban-routing. \cite{joe2020deep} employ a hybrid approach with a genetic algorithm to address a dynamic problem with stochastic customers. Works integrating both stochastic and multi-objective components are rare, at least in routing. \cite{tozer2017many} propose an RL model that selects voting methods based on social choice theory.  \cite{niu2024multi} and \cite{niu2021improved} embed a decision tree that learns node orderings in a genetic algorithm in location and vehicle routing, while \cite{peng2023multi} adapt the same approach but for multi-modal transportation. \cite{zhang2024multimodal}, in contrast, optimize multi-modal transportation routing using a framework aided by simple Q-learning.

Recent innovations in the literature have prompted the use of Deep Reinforcement Learning (DRL) to solve CO problems. Among the earliest of end-to-end ML applications in routing was \cite{kool2018attention} who leverage a Graph Attention network \cite{velivckovic2017graph} to estimate $p_{ij}$. \cite{kwon2020pomo} extended on this by presenting a model that uses multiple solution trajectories during model training and solution inference (generation) which we refer to as Policy Optimization with Multiple Optima (POMO) for simplicity. POMO is able to process complex graph information and greedily decide on the next node visit most likely to minimize the objective value(s). The model is well-credited for its ability to generate solutions of high quality in a relatively short time compared to alternative solvers and heuristics for a multitude of routing problems given a certain distribution. During training, POMO learns the relationship between node orderings and the resulting final objective values to look for orderings that minimize the objective in the future. 

The impressive performance of DRL-based methods like POMO has led to adaptations in multi-objective and stochastic cases for routing. For multi-objective CO routing problems, DRL has been proposed in works like \cite{sarker2020data}, \cite{wang2023multiobjective}, \cite{chen2023efficient} and \cite{lin2022pareto}. For applications in stochastic CO routing problems, works like \cite{schmitt2022learning}, \cite{achamrah2024leveraging}, \cite{iklassov2024reinforcement} and \cite{zhou2023reinforcement} showcase end-to-end methods addressing demand and travel-time uncertainty, which is usually accounted for by fine-tuning the pre-trained model's parameters. In these works, routing problems are solved in an end-to-end fashion whereby the RL model decides on successive node visits. This is in contrast to other DRL innovations that  rely on hybrid methods \citep{jia2025robust,son2024equity} for stochastic and multi-objective applications.

To the best of our knowledge, there is no end-to-end DRL mechanism for solving routing (or CO) problems characterized by both uncertainty and multiple objectives. Many of the multi-objective DRL architectures can not be easily adapted to account for uncertainty and vice-versa. On the other end, existing hybrid methods depend on the heuristics at hand which are often problem-specific. For example, deriving voting methods or learning node orderings with decision trees may not be suitable for more complicated vehicle routing variants such as ones with time windows. Ideally, one would like to come up with an end-to-end model that can be easily adapted to solve a multitude of routing problems without the complexities associated with external assumptions or hyper-heuristics. The novelty of this paper is embedded in deriving such a framework for solving C-VRPTW, although the framework can be easily adopted to solve other routing problems as explained below.

\section{Problem Description}
\label{Problem Des}

Consider a C-VRPTW instance with parameter input $\mathcal{P}$  and stochastic travel times $\mathcal{T}$, where $\mathcal{T} \in \mathcal{P}$. Furthermore, assume the problem is characterized by a set of $K$ objectives for which preference $\lambda_k$ corresponds to objective $f_k(.) \forall k\in K$. The preferences $\lambda_k$ are embedded in a vector $\lambda$. For simplicity, assume $\sum_{k\in K}\lambda_k=1$ and $0\leq f_k(.)$. 

Such preferences could reflect the decision-makers view of importance of one objective relative to another. Such preferences may not be explicitly defined initially, but repeatedly solving the problem with different $\lambda_k$ values enables the decision-maker to understand the relationship between decisions and corresponding objective values through the construction of a Pareto Front. Thereafter, the decision-maker can choose decisions on the Pareto Front that align with their preferences. The proposed model should therefore be able to approximate a Pareto optimal solution for any set of preferences as much as possible.

Let $n$ be the number of nodes that may be visited in a solution and $V$ be the set of nodes, so that $|V|=n$. Each node $i \in V$ is characterized by some features that depend on the problem at hand, such as demand or time-windows. Between each pair of nodes, there is an arc with travel times corresponding to the stochastic parameters $\mathcal{T}$. That said, $\mathcal{T}$ is a $n \times n$ matrix with elements $t_{ij}$ representing the travel times between nodes $i$ and $j \in V$. Let $t_{ij}$ follow some predetermined probability distribution $f_{ij}(.)$. Depending on the problem, a depot may be defined as well. If so, we index the depot by 0, so that $0\in V$ and define stochastic travel times $t_{0i}$ and $t_{i0}$ for each other node $i \in V$.

The objective is to jointly optimize all the $K$ objectives while satisfying operational constraints. In C-VRPTW, time-window constraints, as well as objective values, are largely determined by travel times. As such, we are dealing with a CO problem where the objective values and constraint satisfaction for a set of actions/decisions are uncertain. 

Our proposed model strives to define a mapping - specified by a set of parameters $\theta$ - between graph information and associated node ordering in a solution that maximizes rewards (i.e. delivers good objective values). Our model takes into account the uncertainty in travel times represented by the graph's arc lengths and conflicting rewards (objectives). The mechanism by which we define this mapping serves the purpose of our study. Furthermore, cases with uncertain travel times $\mathcal{T}$ pose a greater challenge than others in the literature such as in \cite{niu2021improved} which deal with stochastic demand. This is because the number of travel time parameters is much more than demand parameters, invoking a larger computational capacity to process the associated uncertainty. Furthermore, different node orderings can give different feasibility and solution scores for a given set of travel time realizations, in contrast to the case with stochastic demand. In the following section, we present a method that treats the problem presented above.

\section{Methodology}
\label{method}

\begin{figure}[H]
\centering
\includegraphics[width=0.9\textwidth]{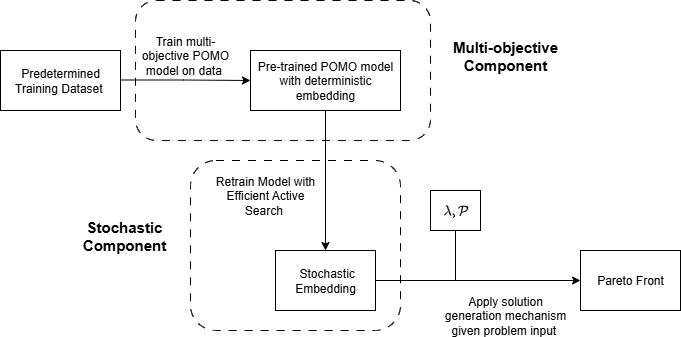}
    \caption{Visual illustration of our complete approach using pre-trained POMO model.}
    \label{Overall Approach}
\end{figure}

We propose a single machine learning model whose training is partitioned into two phases. One phase deals with the multi-objective component of the problem and the latter with the stochastic component. Our model is a variant of the POMO model whose parameters correspond to $\theta$. It is firstly trained on a sample of multi-objective deterministic instances. Afterwards, the parameters embedding information on the travel times are re-trained to account for the stochasticity. The retraining is done by means of an Efficient Active Search (EAS) which we will explain below. Finally, given problem input $\mathcal{P}$ and a vector of preferences $\lambda$, we apply a solution generation procedure with the re-trained POMO model to produce a solution. The Pareto Front can then be estimated by repeatedly solving the problem $\mathcal{P}$ for different values of $\lambda$. An overview of our overall method is illustrated in Figure \ref{Overall Approach}. In the following section, we elaborate on the components of our approach.

\subsection{Multi-Objective Component}
\label{Neural MOCO}

The standard POMO model seen in \cite{kwon2020pomo} defines a mapping with parameters $\theta$ between problem parameter input $\mathcal{P}$ and a resulting reward/objective value. More precisely, the model's main architecture is given by an encoder-decoder duo. The encoder with parameters $\theta_{enc}$ takes as input problem parameters to construct an embedding $\omega$,  which is a summary of the problem features as explained in \cite{kool2018attention}. $\omega$ is then used alongside other dynamic information pertaining to the current status of the solution by the decoder with parameters $\theta_{dec}$ to take actions until a solution is complete with a corresponding reward. Since the reward is constant for every set of parameter input $\mathcal{P}$, the decoder can automatically process $\omega$ and make a decision.

In the multi-objective setting, the reward depends on the preferences $\lambda$. To that end, the information needed to make a decision is not entirely prescribed by the embedding $\omega$. In turn, \cite{lin2022pareto} propose a multi-objective variant of POMO. The idea is to equip the decoder with a Multi-Layer Perceptron with parameters $\psi$ that takes $\lambda$ as input and determines the corresponding decoder parameters $\theta(\lambda|\psi)$. The resulting POMO model can then be used to construct a Pareto Front by repeatedly sampling realizations of $\lambda$ and solving the corresponding problem for a single realization one at a time. Since POMO models can solve a single instance quite efficiently, the Pareto Front would not require a significant computational budget to be constructed.

Figure \ref{Neural MOCO} depicts the process by which a solution is generated with the model of \cite{lin2022pareto}. This model and its training correspond to the Multi-objective component in Figure \ref{Overall Approach}. In the following section, we explain how the embedding $\omega$ generated from this model is updated by means of an EAS to adapt to stochastic travel times $\mathcal{T} \in \mathcal{P}$.

\begin{figure}[H]
\centering
\includegraphics[width=0.9\textwidth]{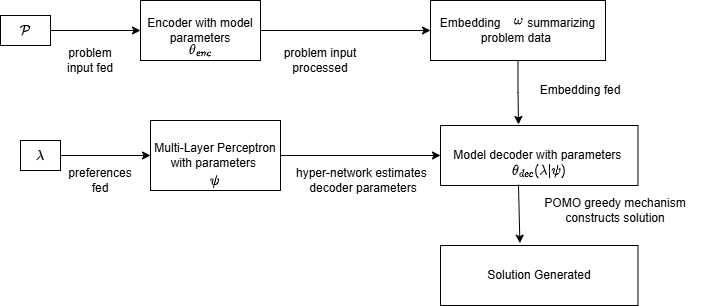}
    \caption{Overview of solution generation with the model of \cite{lin2022pareto}.}
    \label{Neural MOCO}
\end{figure}

To compute the reward for a given solution $\pi_i$ for problem instance $s_i$, \cite{lin2022pareto} propose a linear aggregation technique. The $K$ objectives are aggregated into a single objective (reward) as follows:
\begin{equation}
\label{Tcheby}
    \sum_{k \in K} \lambda_kf_k(\pi_i)
\end{equation}
with $f_k(.)$ representing objective $k$. The term in (\ref{Tcheby}) corresponds to the reward $L(\pi_i|{\lambda, s_i})$ for solution $\pi_i$ given instance $s_i$ and preferences $\lambda$. Solution $\pi_i$ is generated according to a probability distribution $p_{\theta(\lambda)}(\pi_i|s_i)$ which can be constructed in a greedy or sampling fashion.

The model is trained by repeatedly sampling a preference vector $\lambda$ and a corresponding batch of $B$ instances. Thereafter, $M$ different solution trajectories are created where each trajectory represents one possible solution. The average reward over all the $M$ different solution trajectories and $B$ instances is computed, where the reward from one trajectory corresponds to some aggregation of the $K$ objectives such as in (\ref{Tcheby}). The average reward is, in turn, used to compute the loss function and associated gradient for optimizing the model's parameters $\theta$. To estimate the gradient loss, we make use of the REINFORCE training algorithm of \cite{williams1992simple} with the ADAM optimizer. The training algorithm is summarized in Algorithm \ref{Neural MOCO training pseudocode}.

\begin{algorithm}
	\caption{Training POMO Model from \cite{lin2022pareto}}
\begin{algorithmic}[1]
 \State \textbf{Input}: preference distribution $\Lambda$, instances distribution $\mathcal{S}$, number of Training steps $T$, number of objectives $K$, batch size $B$, number of solution trajectories $M$.
 \State{\textbf{Output}: model parameter $\theta$.}
 \For {$t=1\cdots T$}
 \State Sample $\lambda_k$ from $\Lambda$ for $k \in K$.
 \State{Sample $B$ instances from $\mathcal{S}$}
 \State {Generate $M$ different solutions using $p_{\theta(\lambda_k)}(.|s_i)$ for each instance $s_i$.}
 \State{Define shared baseline reward for instance $s_i$ using $b_i=\frac{1}{M}\sum_{j=1}^M L(\pi^j_i|{\lambda, s_i})\hspace{0.75cm} \forall i \{1,\cdots,B\}$ }
 \State{Compute gradient  $\Delta_{\theta}= \frac{1}{BM}\sum_{j=1}^B\sum_{i=1}^M[(L(\pi^j_i|{\lambda, s_i})-b_i)\Delta_{\theta(\lambda)}\log p_{\theta(\lambda)}(\pi^j_i|s_i)$]}
\State{$\theta = ADAM(\theta, \Delta_{\theta})$}
 \EndFor
\State{\textbf{return} $\theta$.}
  \end{algorithmic}
  \label{Neural MOCO training pseudocode}
\end{algorithm}

\subsection{Stochastic Component}
\label{stoch comp}

Given that travel times are not encoded with the problem input for POMO, but implicitly represented by a parametrized embedding $\omega$, an EAS is introduced in \cite{schmitt2022learning} by which the model is re-trained to adapt to the different possible stochastic realizations of $\mathcal{T}$. In the retraining phase, only $\omega$ is updated. In each step, a gradient $\Delta_{\omega}$ is computed with $M$ solution trajectories by the following formula:
\begin{equation}
\label{omega gradient}
  \Delta_{\omega}= \frac{1}{M}\sum_{j=1}^M[(L(\pi^j|{\lambda, s})-b)\Delta_{\omega(\lambda)}\log p_{\omega(\lambda)}(\pi^j|s)]
\end{equation}
and $\omega$ is updated by means of gradient ascent. For consistency, we use the ADAM optimizer for the update. Observe that the gradient computation in (\ref{omega gradient}) is similar to the one in Line 8 in Algorithm \ref{Neural MOCO training pseudocode} - based on the REINFORCE algorithm. The major differences is that this update concerns only one instance rather than a batch of $B$ instances, and the only parameters being updated are $\omega$. As such, the probability distribution $p_{\omega(\lambda)}(\pi^j|s)$ is determined by updates in $\omega$. 

The embedding is updated over a series of $T_\omega$ steps for instance $s$ with expected travel times $E[\mathcal{T}]$ and then evaluated on $W$ stochastic realizations every $t_e$ iterations. $t_e$ can be interpreted as the evaluation frequency. Further, the number $W$ should be sufficiently large to address many possible realizations. Since this evaluation is expensive, it is only done in a few of the $T_\omega$ steps. The $\omega$ value that gives the lowest mean aggregate objective value over all $W$ scenarios in the evaluation steps is selected as the final embedding.

There is an important consideration regarding the active search described in \cite{schmitt2022learning}. In the evaluation epoch, the resulting embedding $\omega$ is evaluated on a large number of realizations $W$. Many of these realizations are similar and the resulting objective values as specified by the pretrained POMO model from Section \ref{Neural MOCO}, rendering these evaluations somewhat repetitive and unnecessary. Ideally, we are interested in evaluations that are distinctive in input and resulting objective values. Secondly, because $W$ is rather large, this constrains the number of evaluations we can carry out during retraining. That said, a more efficient evaluation method may not only reduce retraining time but also allow us to evaluate the embedding more frequently with a smaller $t_e$, giving us a larger search space that may possibly result in a larger embedding.

To account for the aforementioned considerations, we propose some adjustments to the EAS. Firstly, we sample the $W$ possible scenarios. Thereafter, we cluster these methods to get a subset of scenarios that are sufficiently representative of the spectrum of realizations. To do that, we use the clustering method of \cite{abouelrous2022optimizing}. 

We consider the initial embedding $\omega$ generated by training the entire model in Algorithm \ref{Neural MOCO training pseudocode}. We solve, the $W$ problems in batches of $B$ to speed up their evaluation. For each instance $s_i$, we select the solution trajectory from the $M$ trajectories with the highest aggregate reward $L(\pi^*|s_i)= \max_{j\in\{1,\cdots,M\}}L(\pi^j|s_i)$ which will be used to compare with other scenarios. The highest is chosen instead of the mean as it represents the best solution found. The aggregate reward is calculated with fixed preferences $\hat{\lambda}$ during clustering, although the preferences may change later during the active search. Fixing parameter values and assuming perfect information has been shown to be very beneficial for clustering \citep{abouelrous2022optimizing}. Two scenarios $s_i$ and $s_k$ are clustered together if:
\begin{equation}
\label{cluster func}
    |L(\pi^*|s_i)-L(\pi^*|s_k)|<\epsilon
\end{equation}
with $|.|$ being the absolute operator and $\epsilon$ being some threshold. Thus, scenarios $s_i$ and $s_k$ are clustered together if their objective values with the initial embedding are sufficiently close. 

The first scenario automatically forms a cluster. The scenarios are compared to the clusters in order. So, scenario $s_k$ is compared to the scenario representing cluster 1. If it meets (\ref{cluster func}), they are clustered together.  Otherwise, $s_k$ is compared to the scenario in cluster 2 and so forth. If $s_K$ is compared with all existing clusters and not clustered with any, it forms a new cluster. The result of the clustering procedure is a subset of $\overline{W} < W$ scenarios. These scenarios are then used in the embedding evaluations which happen every $t_e$ iterations. Since $\overline{W}$ is much smaller than $W$, we can opt for a smaller $t_e$ value and evaluate the embeddings more frequently and efficiently. 

Our active search is summarized in Algorithm \ref{EAS pseudocode}. An important distinction with the active search proposed in \cite{schmitt2022learning} is that preferences $\lambda$ are repeatedly sampled during re-training to ensure that the uncertainty is incorporated under different sets of preferences determined by the decision-maker. When new preferences are sampled, the decoder parameters have to be updated, but one should ensure that the embedding $\omega$ is not altered as it is being retrained (Line 14). Similarly, during evaluation (Line 20), all decoder parameters are updated except $\omega$. We also fix the preference vector $\lambda$ during evaluation (Line 20) to ensure a fair evaluation as the model's performance for different values of $\lambda$ may be uneven.

The returned $\omega^*$ is then saved to solve the problem instance with unknown travel times using a greedy policy that maximizes the probability of the next visited node. The distribution of nodes to be visited is defined by the retrained POMO model. We refer to the model trained using Algorithm \ref{EAS pseudocode} as \textbf{EAS-cluster}.

\begin{algorithm}[H]
	\caption{EAS-cluster Training}
\begin{algorithmic}[1]
 \State \textbf{Input}: number of search steps $T_{\omega}$, total number of realizations to consider $W$, clustering threshold $\epsilon$, batch size during evaluation $H$, initial preferences $\hat{\lambda}$, evaluation frequency $t_e$.
 \State{\textbf{Output}: updated embedding $\omega^*$}
 \State{\textbf{Initialization}: initial embedding $\omega$}
 \For{ the $W$ scenarios}
 \State{Sample them in batches of $H$}
 \State{Evaluate the $H$ instances using $\omega$ and fixed preferences $\hat{\lambda}$.}
 \State{Cluster the instances according to (\ref{cluster func})}
 \EndFor
 \State{$\overline{W}$ = Nr. of clusters.}
 \State{$\omega^*=\omega$}
 \State{Evaluate $\omega^*$ on the $\overline{W}$ scenarios with $\hat{\lambda}$ to determine initial mean aggregate reward  $L(\pi^*|\overline{W},\omega^*)$ }
 \For{$t=1\cdots T_{\omega}$}
 \State{Sample new preferences $\lambda$}
 \State{Update Decoder parameters without updating $\omega$}
 \State{Solve deterministic instance using current embedding $\omega^*$ for given $\lambda$.}
 \State{Compute gradient $\Delta_{\omega}$ using (\ref{omega gradient}).}
 \State{$\omega=ADAM(\Delta_{\omega},\omega)$}
 \If{$t$ is an evaluation epoch (a multiple of $t_e$)}
  \State{Evaluate $\omega$ on the $\overline{W}$ scenarios with $\hat{\lambda}$. Let $L(\pi^*|\overline{W}, \omega)$ be the resulting mean aggregate reward. Ensure that $\omega$ is not updated in the evaluation.};
 \If{$L(\pi^*|\overline{W}, \omega)<L(\pi^*|\overline{W},\omega^*)$}
 \State{$\omega^*=\omega$}
\State{$L(\pi^*|\overline{W},\omega^*)=L(\pi^*|\overline{W}, \omega)$}
 \EndIf
 \EndIf
 
\EndFor
\State{\textbf{return} $\omega^*$.}
\end{algorithmic}
\label{EAS pseudocode}
\end{algorithm}

\subsection{Extension to other DRL Architectures}

Our proposed framework is largely based on the POMO architecture common to both the multi-objective and stochastic component. In that sense, our framework is largely restricted to POMO-based architectures. However, the architecture is largely considered a pillar in using DRL for solving CO problems \citep{wang2024solving}. Much of the recent literature on multi-objective DRL work mentioned in Section \ref{previous work} also makes use of encoder-decoder architecture of \cite{lin2022pareto} making the framework relevant to many applications.

Extension to other DRL architectures is not straightforward. Assuming another encoder-decoder architecture where the decoder parameters contain a graph embedding $\omega$, the general framework may be still applicable. However, there is no clear guarantee on the performance of the model which is also subject to how the decoder interacts with the preference vector $\lambda$.

Figure \ref{Overall Approach} offers an important guideline, however, into solving stochastic multi-objective problems. One should first establish the multi-objective component before proceeding with the stochastic one. The reason being that the model should first be able to generate multiple diverse solutions for a given problem before being able to assess these solutions on a set of stochastic scenarios. This is particularly evident in the retraining mechanism in Section \ref{stoch comp} whereby preference vectors $\lambda$ are repeatedly sampled during the EAS in Algorithm \ref{EAS pseudocode}.

\section{Numerical Experiments}
\label{Numerical Experiments}

To evaluate our method, we propose a series of numerical experiments where we solve a bi-objective ($K=2$) C-VRPTW with stochastic travel times. We jointly minimize the objective of total travel times and makespan (total time consumed by longest route). We compare the performance of our method on unseen instances relative to alternative methods. For the comparison, we consider the final objective value as well as the total run time. In the following sections, we elaborate on the set-up of our numerical experiments and analyze their results.

\subsection{Set Up}

We consider three instance classes of sizes $n \in \{50,100,200\}$ with $B=20$ instances for each $n$. For all instance classes, the locations are given in 2-D space where the x-y coordinates are sampled from a square of length 1. The vehicle capacities are 40, 50 and 70 for the three classes in increasing $n$. Node demand is sampled uniformly as integer from the interval $[1,9]$. Time windows are such that the lower time window $tw_{low}$ is sampled uniformly as integer from the interval [0,16], the time window width $tw_{width}$ from the interval [2,8] and the upper time window equal to $\min\{tw_{low}+tw_{width}, tw_{horizon}\}$ with the $tw_{horizon}$ being the planning horizon. Naturally, the depot's time window is $[0,tw_{horizon}]$, and we set $tw_{horizon}$ to 18. Service times are sampled uniformly between [0.2,0.5]. Lastly, the mean travel times $E[\mathcal{T}]$ are equal to the euclidean distance, and the travel times follow a normal distribution with standard deviation $0.2\times E[t_{ij}]$ for arc $(i,j)$.

For each instance size $n$, we train a different POMO model with 200 epochs and 100,000 episodes, giving a total of 20,000,000 instances. The instances were sampled from the aforementioned distribution in batches of 64 for $n=50$ and 32 for $n=100, 200$. Training was conducted on a GPU node with 2 Intel Xeon Platinum 8360Y (\cite{hpc2}) Processors and a NVIDIA A100 Accelerator (\cite{hpc3}). The training times of the multi-objective POMO model were 18, 60 and 144 hours for $n=$50, 100 and 200.

For EAS-cluster,  we initially consider a sample of $W=1,000$ scenarios in batches of $H=64$ from which we decide on the clusters $\overline{W}$. In doing so, we consider initial preferences $\hat{\lambda_1}=\hat{\lambda_2}=0.5$ to assign all objectives equal importance when reducing scenarios. Two scenarios are clustered together if their aggregated objective values $\hat{\lambda_1}f_1(\pi^*)+\hat{\lambda_2}f_2(\pi^*)$ differ by less than $\epsilon=1\%$ where $\pi^*$ is the best solution found by the pre-trained ML model from Section \ref{Neural MOCO} before EAS for the corresponding instance. 

For $n=$50, 100 and 200, EAS-cluster produced 7.3, 6.5 and 5.75 clusters on average across the $B=20$ instances. This shows that we can significantly reduce the number of scenarios as many of these correspond to similar figures of $f_1(.)$ and $f_2(.)$ for fixed $\hat{\lambda}$. Furthermore, the number of clusters deceases as $n$ increases, likely due to differences in objective values becoming less significant as objective values like travel time become larger for larger $n$. 

Once the scenarios are clustered, we conduct an active search with $T_{\omega}=2,500$ steps where the embedding is evaluated every $t_e=100$ steps. We use three benchmark methods, that test different aspects of our study. They are as follows:
\begin{itemize}
    \item \textbf{NoEAS:} the model without active search, which is simply the multi-objective POMO model from \cite{lin2022pareto} without any updates to embedding $\omega$ applied to solve the deterministic instance. This baseline has been repeatedly shown to outperform other methods such as NSGA-II \citep{kalyanmoy2002fast} and MOEA/D \cite{zhang2007moea} in solving multi-objective combinatorial optimization problems, especially in routing, so we focus on comparing with it. We generate 101 values of $\lambda$, where $\lambda_1$ is computed from 100 evenly spaced intervals in the range [0,1] and $\lambda_2=1-\lambda_1$ to represent the Pareto Front.
    \item \textbf{EAS-basic:} with the active search of \cite{schmitt2022learning} with $t_e=250$. Similarly, we use the same 100 evenly spaced $\lambda$ realizations as above. Solution inference is done using a greedy policy defined by the retrained model.
    \item  \textbf{LGA:} the method of \cite{niu2021improved}. It is a learning-based Genetic Algorithm which makes use of a decision-tree to learn optimal customer orderings. We adjust this method slightly to allow for time-window configurations, since it was originally developed for a VRP variant with stochastic demand only. Adding time-windows increases the frequency of infeasible solutions. We respond to this by increasing the number of Genetic Algorithm iterations to 2,000 - from the original 200 - and the number of evaluation scenarios in the population to 50 - from the original 10. LGA required no pre-training, although a new ML supervised model had to be trained for each instance during solution generation.
\end{itemize}

The resulting solutions for each method and preference $\lambda$ are evaluated on $R=500$ stochastic realizations of the travel times. The resulting mean objective value is used for comparison as it represents the expected objective value at the Pareto Front for the given $\lambda$. Infeasible realizations are excluded from the evaluation. Solution Inference was done on the same GPU nodes used for training, \citep{hpc2,hpc3}. For LGA, no GPU is needed, so an AMD EPYC 9654 \citep{hpc} CPU node was used. The results of the baselines compared to our method on the proposed set of instances are given in the following section.

\subsection{Results}
\label{results}

To compare results, we consider the same hyper-volume technique mentioned in \cite{lin2022pareto}. Let $\mathcal{P}$ define the Pareto Front for a certain policy and $r^*$ be some reference point that is dominated by all solutions in $\mathcal{P}$. Then, the hyper-volume $HV(\mathcal{P})$ of $\mathcal{P}$ is given by the volume of area $\mathcal{S}$ that is defined as follows:
\begin{equation}
    \mathcal{S} = \{r \in \mathbb{R}^k| \exists y \in \mathcal{P} \text{ such that } y\prec r \prec r^*\}
\end{equation}
where $y \prec r$ indicates that solution $y$ dominates solution $r$. Given a baseline $l$ and instance $b$, the associated hyper-volume is given by $HV(\mathcal{P}^b)_l$. We are particularly interested in the percentage increase in hyper-volume due to using EAS-cluster. This percentage is then averaged over all $B$ instances in the test set to give our main performance measure $Z$ that is given by:
\begin{equation}
    Z=\frac{1}{B} \sum_{b=1}^B \frac{[HV(\mathcal{P}^b)_{EAS-cluster} - HV(\mathcal{P}^b)_l]\times100}{HV(\mathcal{P}^b)_l}
\end{equation}

For EAS-cluster and EAS-basic the retraining times of the active search are compared. We report the average ratio $tf$ of EAS-cluster's active search $t_{av}^{EAS-cluster}$ relative to the active search of EAS-basic $t_{av}^{basic}$. More precisely:
\begin{equation}
    tf_{av} =\frac{t_{av}^{EAS-cluster}}{t_{av}^{EAS-basic}}
\end{equation}
such that values $<1$ indicate a smaller run time with EAS-cluster. Furthermore, we report the total run time for solution inference per instance as $t_{inf}$. For EAS-cluster and EAS-basic, this includes the total run time with the active search and solution generation using POMO's greedy policy for all $\lambda$ values, while only the latter is incorporated for NoEAS which lacks active search. For each of the POMO-based methods, inference times are consistent among all $B$ instances for fixed $n$. For LGA, this includes the solution generation and training of the supervised learning model which happens during inference.

We also report the number of solutions $Q_r$ generated at the Pareto Front by each method. Note that this number is equal for all POMO-based models (101). Lastly, we report the average number of feasible realizations $R^f$ out of the total $R=500$ for the solution constructed from the greedy policy for each $n$. Table \ref{experimental results 1} compares the result of EAS-cluster with the three baselines. Additional statistics on the percentage increase in hyper-volume are reported in Table \ref{add stats 1} in \ref{HV add stats}.

\begin{table}[H]
    \centering
    \begin{tabular}{cc|c|c|c|c}
$\mathbf{n}$&\textbf{Metric}&\textbf{EAS-cluster}&\textbf{EAS-basic}&\textbf{NoEAS}&\textbf{LGA}\\
    \hline
    $50$& $Z$ (avg. ratio)&0&-0.02\%&8.14\%&327.87\%\\
    &$tf_{av}$ (ratio)&1&0.44&$- -$&$- -$\\
    &$t_{inf}$ (mins)&49&58&42&$\approx 0.64$\\
    &$Q_r$ (Nr.)&101&101&101&8\\
    &$R^f$ (Nr./500)&474&494 &494&426\\
    \hline
    $100$&  $Z$ &0&-0.28\%&6.34\%&547.74\%\\
    &$tf_{av}$ &1&0.43&$- -$&$- -$\\
    &$t_{inf}$ &90&105&78&$\approx 0.64$\\
    &$Q_r$ &101&101&101&10\\
    &$R^f$ &438&443&447&380\\
    \hline
    $200$&  $Z$&0&-0.51\%&4.73\%&1,120.21\%\\
    &$tf_{av}$&1&0.40&$- -$&$- -$\\
    &$t_{inf}$&164&195&143&$\approx 0.64$\\
    &$Q_r$&101&101&101&13\\
    &$R^f$&422&430&410&257\\
\end{tabular}
    \caption{Results EAS-cluster compared to EAS-basic on 20 different multi-objective C-VRPTW instances with stochastic travel times $\mathcal{T}$ for each instance size $n$.}
    \label{experimental results 1}
\end{table}

For EAS-cluster, 7, 12 and 21 minutes per instance were spent on the EAS alone for the given values of $n$. For all values of $n$ (-0.02\%,-0.28\% and -0.51\%), EAS-cluster results in a slightly worse Pareto Front than EAS-basic as given by values of $Z$ close to zeros. This slight deterioration could be due to the bias imposed by the evaluation on the few scenarios in $\overline{W}$. To emphasize on the distribution of the value of $Z$ among the $B=20$ instances, we provide Figure \ref{z hist tf} in \ref{Figure app}. EAS-basic incurs a larger computational during the active search, consuming more than twice the run time of EAS-cluster for all instances as result of the larger number of evaluations on scenarios sampled from $W$. This gives $tf_{av}$ ratios of 0.44, 0.43 and 0.40 for the corresponding values of $n$, despite EAS-cluster incurring more than twice the evaluations of EAS-basic (25 to 10) during the 2,500 retraining steps. The resulting run times for EAS-basic are 58, 105 and 195 minutes per instance compared to EAS-cluster's 49, 90 and 164 minutes. Feasibility also slightly improves with EAS-basic as the number of feasible evaluations $R^f$ is 494 443 and 430 compared to EAS cluster's 474, 438 and 422 which could be attributed to the aforementioned bias. With these results, we observe that the reduction in computation time due to EAS-cluster outweighs the small deterioration in the Pareto Front hyper-volume and scenario feasibility.

The added value of active search, and EAS-cluster specifically, is shown in the significant improvement in hypervolume compared to NoEAS, with $Z$ values of 8.14\%, 6.34\% and 4.73\% for the given $n$.  For the sake of clarity, we provide the distribution of the $Z$ values in Figure \ref{z hist ff} in \ref{Figure app}. We see that $Z$ decreases as $n$ increases due to the increased difficulty of solving larger problems. The run times of NoEAS only include the time for Pareto Front construction which is similar for EAS-cluster, giving 42, 78 and 143 minutes. Solution feasibility is better for $n=50$ and 100, with $R^f=494$ and 447 scenarios. However, for $n=200$, $R^f$ decreases to 410. This could be possibly due to the more pronounced effect of stochasticity in larger instances that can not be easily treated by the original embedding $\omega$. In fact, $R^f$ decreases as $n$ increases for all methods as it becomes increasingly difficult for a given solution to meet all the time windows of the $n$ customers.

Finally, we consider $LGA$. Since the existing implementation we found makes use of multi-threading, it was difficult to accurately estimate the run time per instance, which we observed to be less than a minute. However, LGA gives significantly inferior solutions compared to EAS-cluster as given by $Z$ values of 327.87\%, 547.74\% and 1,120.21\%. This can largely be attributed to the methods inability to incorporate information on time-windows in the decision tree's leaves through simple customer orderings, thus struggling to find quality feasible solutions. A prominent advantage of POMO-based models, on the other hand, is their ability to learn complex CO problems which can be easily configured in the model's environment and adequately processed by the encoder. LGA is also unable to find as many solutions as POMO-based methods on the Pareto Front with 8, 10 and 13 solutions on average with respect to the 101 of the POMO-based models. Furthermore, since these solution do not correspond to a specific balance of objectives $f_k(.)$, it is difficult to verify their diversity for different decision-making criteria. In contrast, POMO-based methods can provide such diversity through $\lambda$ parameters to balance the objectives $f_k(.)$.

For a clear comparison of the resulting Pareto Front of all the methods, we refer to Figure \ref{Pareto compare} in \ref{Figure app}. The figure compares the Pareto Front from EAS-cluster with each of the baselines for one of the test instances of size 200. The comparison with EAS-basic in Figure \ref{pareto cluster vs basic} shows that the Pareto Fronts are fairly close, while the comparison with NoEAS in Figure \ref{pareto cluster vs noEAS} shows a more prominent difference especially when preferences are more oriented towards minimizing travel-times. Lastly, the comparison with LGA in Figure \ref{pareto cluster vs LGA} demonstrates the inferiority of LGA's Pareto Front given its inability to accurately encode the problem's parameters.

\subsection{Ablation Studies}
\label{Ablation}

To study the ability of our method to deliver improved results and generalize to settings with different distributions, we propose two sets of experiments. In the first set of experiments, we combine our method with Monte Carlo Simulation. This is in line with the recommendations in \cite{schmitt2022learning} who state that Monte Carlo Simulation may improve over a greedy policy. In the second set of experiments, we apply our method to solve a series of instances from a different distribution with more variable travel times.

\subsubsection{Monte Carlo Simulation}
\label{MS sim}

With the final embedding $\omega^*$, \cite{schmitt2022learning} use Monte Carlo simulation to evaluate the different actions at every step of the solution generation, i.e. node visit. More precisely, at each step, the top 5 recommended actions are evaluated by applying them in the succeeding step and completing each of the 5 solution trajectories using the trained model's policy. For each step, the process is repeated for a fixed number of runs $ms_r$ and the average is taken to determine the best action in the next step.

We make use of the same Monte Carlo simulation technique. There are some adjustments, however, we ought to incorporate in the application of the Monte Carlo Simulation for VRP. Firstly, \cite{schmitt2022learning} consider a large number of simulation runs $ms_r$ - going up to 100s - for each problem instance. This results in long running times per solution as demonstrated by their numerical experiments with instances of size 200 going up to 48 minutes. Since we repeatedly have to apply this procedure for several realizations of $\lambda$ as each realization of $\lambda$ invokes a different solution, we consider a smaller value $ms_r=10$ so that the total run time per instance falls within the same order of magnitude of around 1 hour. Furthermore, we limit the number of sampled $\lambda$ values to 11 instead of 101. 

Unlike the orienteering problem in \cite{schmitt2022learning}, it is not possible to realize the stochastic travel times after taking a step since repeated visits to the depot in VRP would then imply that we can travel back in time. As such, the travel time realization we make after taking a decision concerns the expected value of the travel time, while stochasticity is incorporated in future decisions by the Monte Carlo runs. To impose feasibility constraints we require that the decisions considered at each step are feasible for at least half of the $ms_r=10$ runs, otherwise they are discarded. We require that solutions are only feasible $50\%$ of the time. Similar to the greedy policy, we evaluate the resulting solution on a sample of $R=500$ realizations and use the average to estimate the objective values for given $\lambda$ at the Pareto Front.

We compare the performance of the Monte Carlo approach to the greedy policy of EAS-cluster from Section \ref{results} in Table \ref{experimental results 2}. Additional statistics on the percentage increase in hyper-volume are reported in Table \ref{add stats 2} in \ref{HV add stats}. Due to the expensive evaluation, we only consider $n=$50 and 100. The reported $t_{inf}$ times also include the run times for active search which is identical for both methods as they used the same embedding.

 \begin{table}[H]
    \centering
    \begin{tabular}{cc|c|c}
$\mathbf{n}$&\textbf{Metric}&\textbf{EAS-cluster (Monte Carlo Simulation) }&\textbf{EAS-cluster (Greedy)}\\
    \hline
    $50$& $Z$ (avg. ratio)&0&0.33\%\\
    &$t_{inf}$ (mins)&17&49\\
    &$Q_r$ (Nr.)&11&101\\
    &$R^f$ (Nr./500)&489&474\\
    \hline
    $100$&  $Z$&0&-1.26\%\\
    &$t_{inf}$&37 &90\\
    &$Q_r$&11&101\\
    &$R^f$&470&438\\
\end{tabular}
    \caption{Results of EAS-cluster with Monte Carlo Simulation compared to greedy policy for each instance size $n$.}
    \label{experimental results 2}
\end{table}

The results generally imply that monte carlo simulation is of limited added value to multi-objective optimization. While a marginal improvement of 0.33\% is observed for $n=50$, a deterioration of 1.26\% is observed for $n=100$ compared to using the greedy policy. The average run time per instance as given by $t_{inf}$ is slightly less than a third of the greedy policy's run time, with Monte Carlo simulation requiring on average 17 and 37 minutes per instance for $n=50$ and 100. Yet, the number of solutions on the Pareto Front generated from the greedy policy is 10 times as much.

Nonetheless, the Monte Carlo simulation slightly improves the number of feasible realizations from 474 to 489 and from 438 to 470 for $n=$50 and 100. This could be attributed to the repetitive evaluations in the Monte Carlo runs that require that a node visit be feasible for at least half the $ms_r=10$ for it to be considered in the solution. As a result, a stronger feasibility requirement is imposed compared to the greedy policy and more scenarios are satisfied by the resulting solution. For cases where feasibility is crucial, one might still resort to using Monte Carlo Simulation rather than a greedy policy.

\subsubsection{Travel Time Distribution}
\label{TT dist}

In this section, we study the performance of our retrained model EAS-cluster on a dataset with a different travel time distribution. We refrain from conducting an active search and simply run the greedy policy on the new dataset to test model generalization. For the new dataset, we consider standard deviation $E[t_{ij}]\times 0.4$ which is twice as variable as the travel times considered in Section \ref{results}. 

As a benchmark, we consider NoEAS again. The rationale being that if the difference of $Z$ is still positive, then the model with updated embedding $\omega^*$ generalizes well to a reasonable extent beyond a model where no retraining is involved. We consider the greedy policy with $Q_r=101$ solutions again. Since solution inference times $t_{inf}$ and number of Pareto Front points $Q_r$ are identical, we only report $Z$ and $R^f$ in Table \ref{experimental results 3}. Additional statistics on the percentage increase in hyper-volume are reported in Table \ref{add stats 3} in \ref{HV add stats}.

 \begin{table}[H]
    \centering
    \begin{tabular}{cc|c|c}
$\mathbf{n}$&\textbf{Metric}&\textbf{EAS-cluster (Greedy)}&\textbf{NoEAS}\\
    \hline
    $50$& $Z$ (avg. ratio)&0&8.39\% \\
    &$R^f$ (Nr./500)&448&480\\
    \hline
    $100$&  $Z$&0&6.56\%\\
    &$R^f$&391&413\\
    \hline
    $200$&  $Z$&0&4.35\%\\
    &$R^f$&334&340\\
\end{tabular}
    \caption{Results of EAS-cluster with greedy policy compared to NoEAS and more variable travel times for each instance size $n$.}
    \label{experimental results 3}
\end{table}

The results are largely in line with those in Section \ref{results}. EAS-cluster improves the hyper-volume by 8.39\%, 6.56\% and 4.35\% on average for $n=50,$ 100 and 200. While the performance of EAS-cluster may deteriorate slightly in a more variable dataset, it still outperforms NoEAS whose embedding $\omega$, is significantly `out-of-tune' with the more variable travel times $\mathcal{T}$. NoEAS, however, provides more feasible solutions, giving $R^f=$ 480, 413 and 340 solutions compared to EAS-cluster's 448, 391 and 334 for given $n$. This could be explained by the increased bias of cluster evaluations in a different dataset, although the difference in $R^f$ values decreases with larger $n$ due to the increased difficulty of solving larger instances as explained in Section \ref{results}.

\section{Conclusions}
 \label{Conclusionss}

In this paper, we provided an end-to-end method to solve a multi-objective stochastic C-VRPTW. We specifically considered stochastic travel times and minimized total travel distance and make-span. Our model is based on a variant of POMO that is first trained on deterministic multi-objective instance and then retrained through an active search to account for travel time stochasticity. The model could not only be configured to treat a variety of routing problems through its flexible environment, but also account for more different objectives. The end-to-end mechanism of the model not only simplifies solution generation, but is also invariant of other external assumptions on the model environment and the action space.

We provide a training mechanism for the model and enhance it using a clustering algorithm that speeds up evaluations during active search and run time as a result. Our results show that our model is very close in performance compared to a baseline employing standard active search from the literature while being faster. It also significantly outperformed another POMO baseline that disregards stochasticity as well as a common learning-based method for VRP from the literature, resulting in more dominant Pareto Fronts. 

Our model is able to generate as many solutions as desired for any set of preferences and in reasonable run times. Its greedy policy is also sufficient to generate good results, reducing the need for sampling policies often associated with POMO-based models. It also generalizes well to other instances with different distribution than the one encountered during active search. Nonetheless, it often results in less feasible solutions compared to the considered baseline, possibly due to the bias induced by the clustered sample of scenarios.  

Future research should focus more on synchronizing multi-objective training with active search and the derivation of superior clustering techniques by which the associated model bias could be overcome.

\bibliography{MyBIB}

@article{kwon2020pomo,
  title={{POMO}: Policy optimization with multiple optima for reinforcement learning},
  author={Kwon, Yeong-Dae and Choo, Jinho and Kim, Byoungjip and Yoon, Iljoo and Gwon, Youngjune and Min, Seungjai},
  journal={Advances in Neural Information Processing Systems},
  volume={33},
  pages={21188--21198},
  year={2020}
}

@article{kool2018attention,
  title={Attention, learn to solve routing problems!},
  author={Kool, Wouter and Van Hoof, Herke and Welling, Max},
  journal={arXiv preprint arXiv:1803.08475},
  year={2018}
}

@article{velivckovic2017graph,
  title={Graph attention networks},
  author={Veli{\v{c}}kovi{\'c}, Petar and Cucurull, Guillem and Casanova, Arantxa and Romero, Adriana and Lio, Pietro and Bengio, Yoshua},
  journal={arXiv preprint arXiv:1710.10903},
  year={2017}
}

@inproceedings{schmitt2022learning,
  title={Learning to solve a stochastic orienteering problem with time windows},
  author={Schmitt-Ulms, Fynn and Hottung, Andr{\'e} and Sellmann, Meinolf and Tierney, Kevin},
  booktitle={International Conference on Learning and Intelligent Optimization},
  pages={108--122},
  year={2022},
  organization={Springer}
}

@article{bengio2021machine,
  title={Machine learning for combinatorial optimization: a methodological tour d’horizon},
  author={Bengio, Yoshua and Lodi, Andrea and Prouvost, Antoine},
  journal={European Journal of Operational Research},
  volume={290},
  number={2},
  pages={405--421},
  year={2021},
  publisher={Elsevier}
}

@article{giuffrida2022optimization,
  title={Optimization and machine learning applied to last-mile logistics: A review},
  author={Giuffrida, Nadia and Fajardo-Calderin, Jenny and Masegosa, Antonio D and Werner, Frank and Steudter, Margarete and Pilla, Francesco},
  journal={Sustainability},
  volume={14},
  number={9},
  pages={5329},
  year={2022},
  publisher={MDPI}
}

@inproceedings{d2020learning,
  title={Learning 2-opt heuristics for the traveling salesman problem via deep reinforcement learning},
  author={d O Costa, Paulo R and Rhuggenaath, Jason and Zhang, Yingqian and Akcay, Alp},
  booktitle={Asian conference on machine learning},
  pages={465--480},
  year={2020},
  organization={PMLR}
}

@article{mazyavkina2021reinforcement,
  title={Reinforcement learning for combinatorial optimization: A survey},
  author={Mazyavkina, Nina and Sviridov, Sergey and Ivanov, Sergei and Burnaev, Evgeny},
  journal={Computers \& Operations Research},
  volume={134},
  pages={105400},
  year={2021},
  publisher={Elsevier}
}

@article{zhang2021solving,
  title={Solving dynamic traveling salesman problems with deep reinforcement learning},
  author={Zhang, Zizhen and Liu, Hong and Zhou, MengChu and Wang, Jiahai},
  journal={IEEE Transactions on Neural Networks and Learning Systems},
  volume={34},
  number={4},
  pages={2119--2132},
  year={2021},
  publisher={IEEE}
}

@article{accorsi2022guidelines,
  title={Guidelines for the computational testing of machine learning approaches to vehicle routing problems},
  author={Accorsi, Luca and Lodi, Andrea and Vigo, Daniele},
  journal={Operations Research Letters},
  volume={50},
  number={2},
  pages={229--234},
  year={2022},
  publisher={Elsevier}
}

@article{lin2022pareto,
  title={Pareto set learning for neural multi-objective combinatorial optimization},
  author={Lin, Xi and Yang, Zhiyuan and Zhang, Qingfu},
  journal={arXiv preprint arXiv:2203.15386},
  year={2022}
}

@article{abouelrous2022optimizing,
  title={Optimizing the inventory and fulfillment of an omnichannel retailer: a stochastic approach with scenario clustering},
  author={Abouelrous, Abdo and Gabor, Adriana F and Zhang, Yingqian},
  journal={Computers \& Industrial Engineering},
  volume={173},
  pages={108723},
  year={2022},
  publisher={Elsevier}
}

@article{williams1992simple,
  title={Simple statistical gradient-following algorithms for connectionist reinforcement learning},
  author={Williams, Ronald J},
  journal={Machine learning},
  volume={8},
  pages={229--256},
  year={1992},
  publisher={Springer}
}

@misc{hpc2, 
  author = {Intel},
  title={Intel® Xeon® Platinum 8360Y Processor },
  note = {Accessed: 26-02-2025},
  url={https://www.intel.com/content/www/us/en/products/sku/212459/intel-xeon-platinum-8360y-processor-54m-cache-2-40-ghz/specifications.html}, 
  year={2025}}

@misc{hpc3, 
  author = {Nvidia},
  title={{NVIDIA} {A100} Tensor Core {GPU}},
  note = {Accessed: 26-02-2025},
  url={https://www.nvidia.com/en-us/data-center/a100/}, 
  year={2025}}

@article{niu2021improved,
  title={An improved learnable evolution model for solving multi-objective vehicle routing problem with stochastic demand},
  author={Niu, Yunyun and Kong, Detian and Wen, Rong and Cao, Zhiguang and Xiao, Jianhua},
  journal={Knowledge-Based Systems},
  volume={230},
  pages={107378},
  year={2021},
  publisher={Elsevier}
}

@misc{hpc, 
  author = {AMD},
  title={Server Processor Specifications },
  note = {Accessed: 26-02-2025},
  url={https://www.amd.com/en/products/specifications/server-processor.html}, 
  year={2025}}

@article{wu2024multiobjective,
  title={Multiobjective vehicle routing optimization with time windows: A hybrid approach using deep reinforcement learning and nsga-ii},
  author={Wu, Rixin and Wang, Ran and Hao, Jie and Wu, Qiang and Wang, Ping and Niyato, Dusit},
  journal={IEEE Transactions on Intelligent Transportation Systems},
  year={2024},
  publisher={IEEE}
}

@inproceedings{sarker2020data,
  title={A data-driven reinforcement learning based multi-objective route recommendation system},
  author={Sarker, Ankur and Shen, Haiying and Kowsari, Kamran},
  booktitle={2020 IEEE 17th international conference on mobile ad hoc and sensor systems (mass)},
  pages={103--111},
  year={2020},
  organization={IEEE}
}

@article{deng2024multi,
  title={Multi-task multi-objective evolutionary search based on deep reinforcement learning for multi-objective vehicle routing problems with time windows},
  author={Deng, Jianjun and Wang, Junjie and Wang, Xiaojun and Cai, Yiqiao and Liu, Peizhong},
  journal={Symmetry},
  volume={16},
  number={8},
  pages={1030},
  year={2024},
  publisher={MDPI}
}

@inproceedings{yao2017efficient,
  title={An efficient learning-based approach to multi-objective route planning in a smart city},
  author={Yao, Yuan and Peng, Zhe and Xiao, Bin and Guan, Jichang},
  booktitle={2017 IEEE International Conference on Communications (ICC)},
  pages={1--6},
  year={2017},
  organization={IEEE}
}

@inproceedings{iklassov2024reinforcement,
  title={Reinforcement learning for solving stochastic vehicle routing problem},
  author={Iklassov, Zangir and Sobirov, Ikboljon and Solozabal, Ruben and Tak{\'a}{\v{c}}, Martin},
  booktitle={Asian Conference on Machine Learning},
  pages={502--517},
  year={2024},
  organization={PMLR}
}

@article{horvitz2013reasoning,
  title={Reasoning about beliefs and actions under computational resource constraints},
  author={Horvitz, Eric J},
  journal={arXiv preprint arXiv:1304.2759},
  year={2013}
}

@inproceedings{ngatchou2005pareto,
  title={Pareto multi objective optimization},
  author={Ngatchou, Patrick and Zarei, Anahita and El-Sharkawi, A},
  booktitle={Proceedings of the 13th international conference on, intelligent systems application to power systems},
  pages={84--91},
  year={2005},
  organization={IEEE}
}

@article{bayliss2021machine,
  title={Machine learning based simulation optimisation for urban routing problems},
  author={Bayliss, Christopher},
  journal={Applied Soft Computing},
  volume={105},
  pages={107269},
  year={2021},
  publisher={Elsevier}
}

@inproceedings{joe2020deep,
  title={Deep reinforcement learning approach to solve dynamic vehicle routing problem with stochastic customers},
  author={Joe, Waldy and Lau, Hoong Chuin},
  booktitle={Proceedings of the international conference on automated planning and scheduling},
  volume={30},
  pages={394--402},
  year={2020}
}

@article{zhou2023reinforcement,
  title={Reinforcement Learning-based approach for dynamic vehicle routing problem with stochastic demand},
  author={Zhou, Chenhao and Ma, Jingxin and Douge, Louis and Chew, Ek Peng and Lee, Loo Hay},
  journal={Computers \& Industrial Engineering},
  volume={182},
  pages={109443},
  year={2023},
  publisher={Elsevier}
}

@article{niu2024multi,
  title={Multi-objective location-routing optimization based on machine learning for green municipal waste management},
  author={Niu, Yunyun and Xu, Chang and Liao, Shubing and Zhang, Shuai and Xiao, Jianhua},
  journal={Waste Management},
  volume={181},
  pages={157--167},
  year={2024},
  publisher={Elsevier}
}

@article{peng2023multi,
  title={Multi-objective optimization for multimodal transportation routing problem with stochastic transportation time based on data-driven approaches},
  author={Peng, Yong and Gao, Shu Han and Yu, Dennis and Xiao, Yun Peng and Luo, Yi Juan},
  journal={RAIRO-Operations Research},
  volume={57},
  number={4},
  pages={1745--1765},
  year={2023},
  publisher={EDP Sciences}
}

@article{zhang2024multimodal,
  title={Multimodal transportation routing optimization based on multi-objective Q-learning under time uncertainty},
  author={Zhang, Tie and Cheng, Jia and Zou, Yanbiao},
  journal={Complex \& Intelligent Systems},
  volume={10},
  number={2},
  pages={3133--3152},
  year={2024},
  publisher={Springer}
}

@article{kalyanmoy2002fast,
  title={A fast and elitist multi-objective genetic algorithm: NSGA-II},
  author={Kalyanmoy, Deb},
  journal={IEEE Trans. on Evolutionary Computation},
  volume={6},
  number={2},
  pages={182--197},
  year={2002}
}

@article{wang2024solving,
  title={Solving combinatorial optimization problems with deep neural network: A survey},
  author={Wang, Feng and He, Qi and Li, Shicheng},
  journal={Tsinghua Science and Technology},
  volume={29},
  number={5},
  pages={1266--1282},
  year={2024},
  publisher={TUP}
}

@article{zhang2007moea,
  title={MOEA/D: A multiobjective evolutionary algorithm based on decomposition},
  author={Zhang, Qingfu and Li, Hui},
  journal={IEEE Transactions on evolutionary computation},
  volume={11},
  number={6},
  pages={712--731},
  year={2007},
  publisher={IEEE}
}

@article{tozer2017many,
  title={Many-objective stochastic path finding using reinforcement learning},
  author={Tozer, Bentz and Mazzuchi, Thomas and Sarkani, Shahram},
  journal={Expert Systems with Applications},
  volume={72},
  pages={371--382},
  year={2017},
  publisher={Elsevier}
}

@article{wang2023multiobjective,
  title={Multiobjective combinatorial optimization using a single deep reinforcement learning model},
  author={Wang, Zhenkun and Yao, Shunyu and Li, Genghui and Zhang, Qingfu},
  journal={IEEE transactions on cybernetics},
  volume={54},
  number={3},
  pages={1984--1996},
  year={2023},
  publisher={IEEE}
}

@article{chen2023efficient,
  title={Efficient meta neural heuristic for multi-objective combinatorial optimization},
  author={Chen, Jinbiao and Wang, Jiahai and Zhang, Zizhen and Cao, Zhiguang and Ye, Te and Chen, Siyuan},
  journal={Advances in Neural Information Processing Systems},
  volume={36},
  pages={56825--56837},
  year={2023}
}

@article{achamrah2024leveraging,
  title={Leveraging transfer learning in deep reinforcement learning for solving combinatorial optimization problems under uncertainty},
  author={Achamrah, Fatima Ezzahra},
  journal={IEEE Access},
  year={2024},
  publisher={IEEE}
}

@inproceedings{son2024equity,
  title={Equity-transformer: Solving np-hard min-max routing problems as sequential generation with equity context},
  author={Son, Jiwoo and Kim, Minsu and Choi, Sanghyeok and Kim, Hyeonah and Park, Jinkyoo},
  booktitle={Proceedings of the AAAI Conference on Artificial Intelligence},
  volume={38},
  number={18},
  pages={20265--20273},
  year={2024}
}

@inproceedings{jia2025robust,
  title={How Robust Reinforcement Learning Enables Courier-Friendly Route Planning for Last-Mile Delivery?},
  author={Jia, Ziying and Dong, Zeyu and Yin, Miao and He, Sihong},
  booktitle={ICML 2025 Workshop on Programmatic Representations for Agent Learning}
}
\bibliographystyle{model5-names}

\newpage

\appendix

\section{Figures}
\label{Figure app}

\begin{figure}[H]
  \centering
\begin{subfigure}{0.48\textwidth}
\includegraphics[width=0.99\textwidth]{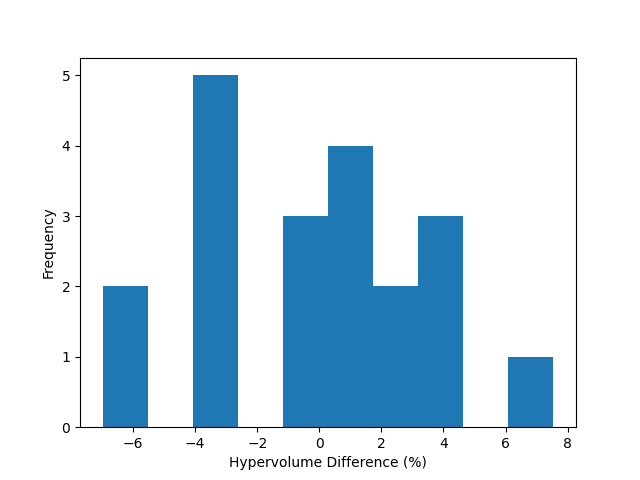}
    \caption{$n=50$}
    \label{z hist 50 tf}
    \end{subfigure}
\\
\begin{subfigure}{0.48\textwidth}
    \includegraphics[width=0.99\textwidth]{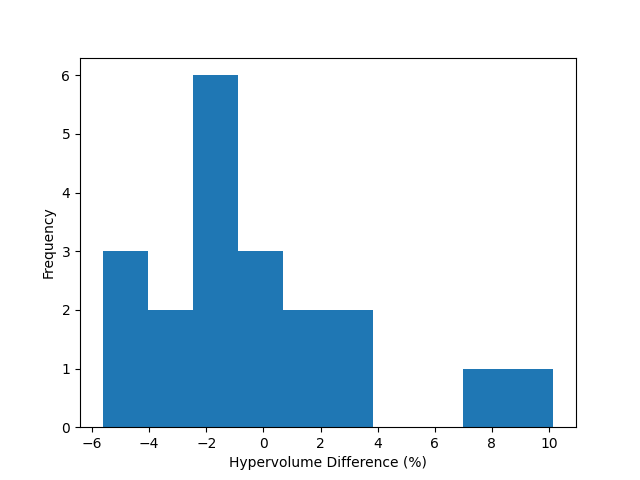}
    \caption{$n=100$}
    \label{z hist 100 tf}
    \end{subfigure}
 \begin{subfigure}{0.48\textwidth}
    \includegraphics[width=0.99\textwidth]{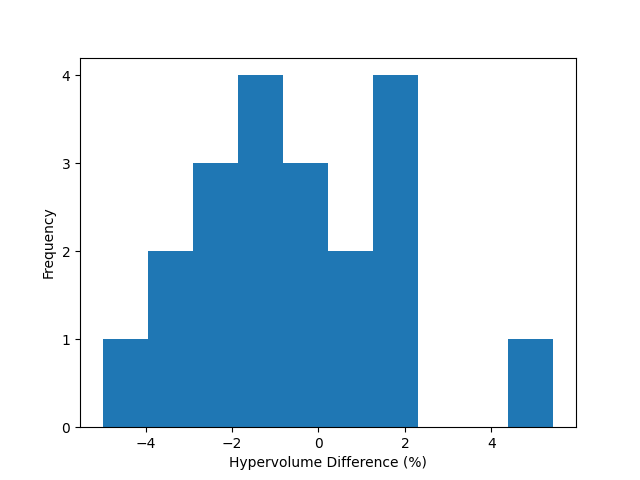}
    \caption{$n=200$}
    \label{z hist 200 tf}
    \end{subfigure}
    \caption{Histogram of $Z$ values of EAS-cluster against EAS-basic.}
    \label{z hist tf}
\end{figure}

\begin{figure}[H]
  \centering
\begin{subfigure}{0.48\textwidth}
\includegraphics[width=0.99\textwidth]{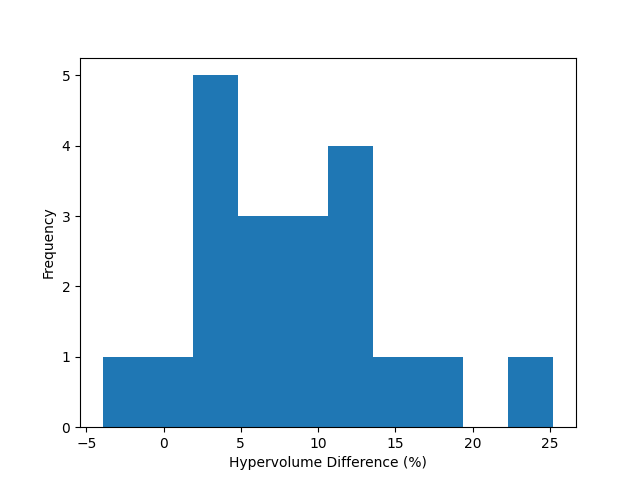}
    \caption{$n=50$}
    \label{z hist 50 ff}
    \end{subfigure}
\\
\begin{subfigure}{0.48\textwidth}
    \includegraphics[width=0.99\textwidth]{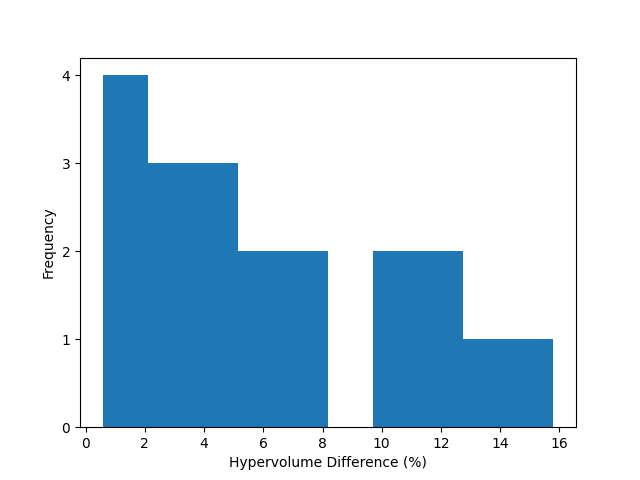}
    \caption{$n=100$}
    \label{z hist 100 ff}
    \end{subfigure}
 \begin{subfigure}{0.48\textwidth}
    \includegraphics[width=0.99\textwidth]{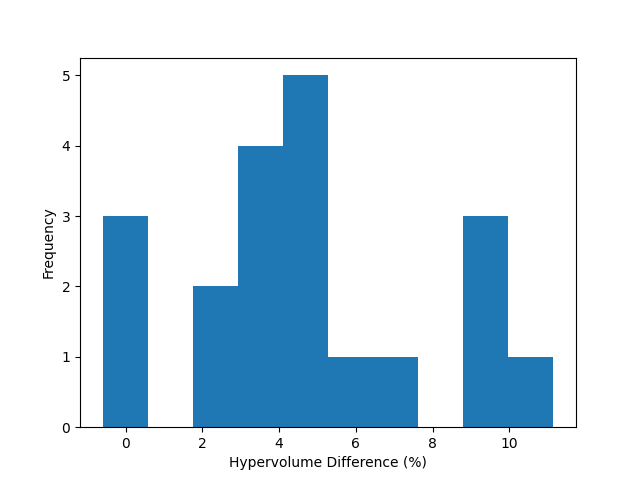}
    \caption{$n=200$}
    \label{z hist 200 ff}
    \end{subfigure}
    \caption{Histogram of $Z$ values of EAS-cluster against NoEAS}
    \label{z hist ff}
\end{figure}

\begin{figure}[H]
  \centering
\begin{subfigure}{0.48\textwidth}
\includegraphics[width=0.99\textwidth]{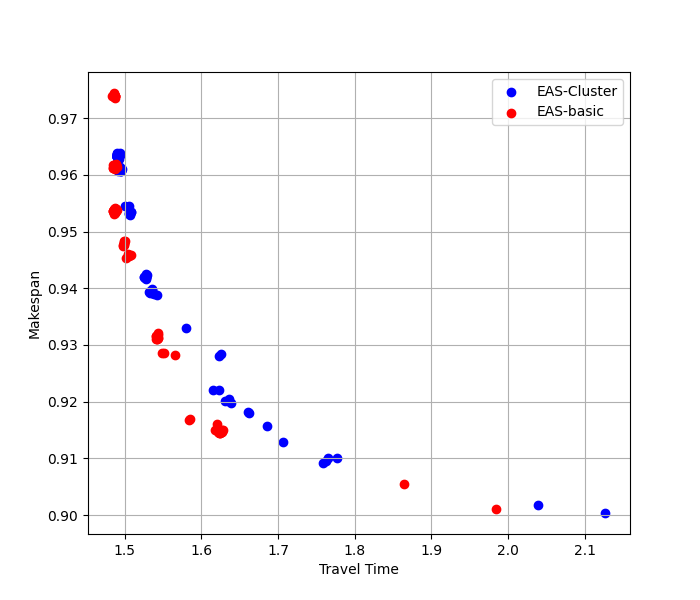}
    \caption{EAS-cluster vs. EAS-basic}
    \label{pareto cluster vs basic}
    \end{subfigure}
\\
\begin{subfigure}{0.48\textwidth}
    \includegraphics[width=0.99\textwidth]{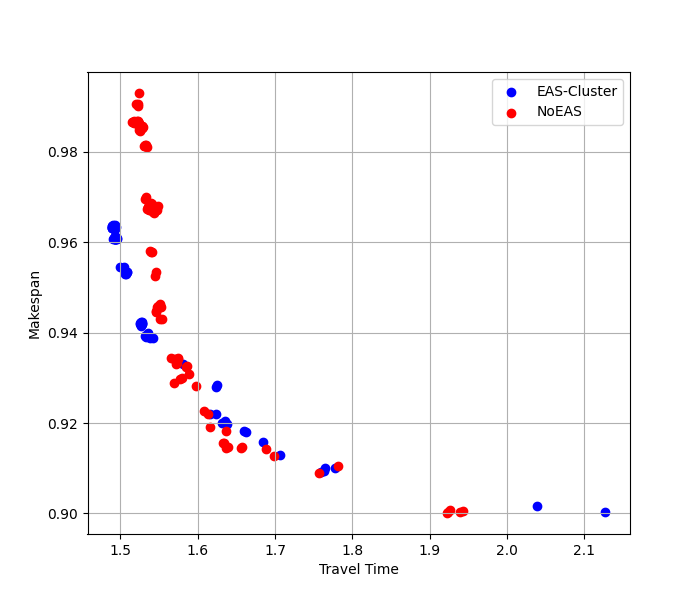}
    \caption{EAS-cluster vs. No EAS}
    \label{pareto cluster vs noEAS}
    \end{subfigure}
 \begin{subfigure}{0.48\textwidth}
    \includegraphics[width=0.99\textwidth]{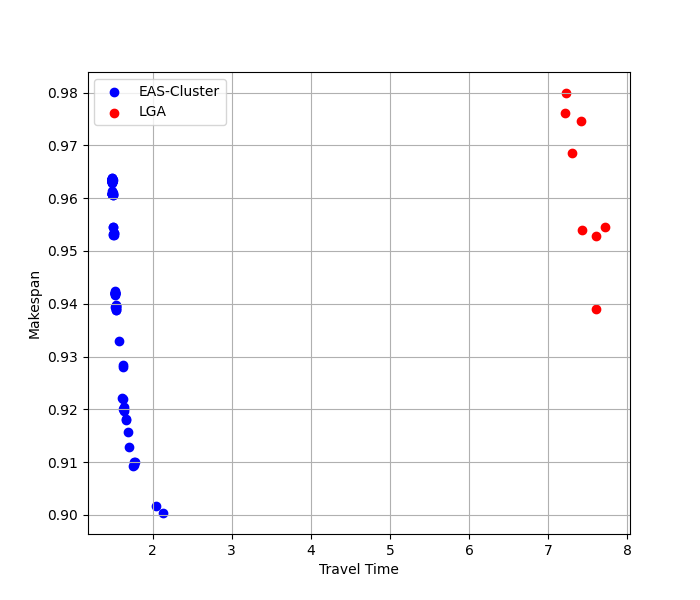}
    \caption{EAS-cluster vs. LGA}
    \label{pareto cluster vs LGA}
    \end{subfigure}
    \caption{Plots comparing the Pareto Front generated by EAS-cluster against other methods for one of the instances of size 200.}
    \label{Pareto compare}
\end{figure}

\section{Hyper-volume Additional Statistics}
\label{HV add stats}

In this section, we report additional statistics on the percentage increase in hyper-volume for our experiments. Our statistics concern the quantity:
\begin{equation}
    \frac{[HV(\mathcal{P}^b)_{EAS-cluster} - HV(\mathcal{P}^b)_l]\times100}{HV(\mathcal{P}^b)_l}
\end{equation}
for which the mean across all $B$ instances represents the metric $Z$ used in Section \ref{results}. In addition to the mean, we report the standard deviation, the min and max values in the following Tables. Table \ref{add stats 1} concerns results in Table \ref{experimental results 1}, Table \ref{add stats 2} concerns results in Table \ref{experimental results 2} and Table \ref{add stats 3} concerns results in Table \ref{experimental results 3}.

\begin{table}[H]
    \centering
    \begin{tabular}{cc|c|c|c}
$\mathbf{n}$&\textbf{Metric}&\textbf{EAS-basic}&\textbf{NoEAS}&\textbf{LGA}\\
    \hline
    $50$& St Dev &3.72\%&6.4\%&97.23\%\\
    & Max&7.54\%&25.23\%&605.76\% \\
    & Min&-6.98\%&-3.94\%& 184.69\%\\
    \hline
    $100$&  St Dev & 3.97 \%&4.63 \%&86.32\%\\
    & Max& 10.16 \% & 15.8 \%&663.51 \%\\ 
    & Min&-5.63 \%&0.57 \%&380.77 \%\\
    \hline
    $200$&  St Dev & 2.37 \%& 3.34 \%&307.28 \%\\
    & Max&5.44 \% & 11.15 \%& 2092.01 \%\\
    & Min&-5.0 \%& -0.6 \%& 758.97 \%\\
\end{tabular}
    \caption{Statistics on percentage increase in hyper-volume of EAS-cluster relative to baseline for the experiments reported in Table \ref{experimental results 1}}
    \label{add stats 1}
\end{table}

\begin{table}[H]
    \centering
    \begin{tabular}{cc|c}
$\mathbf{n}$&\textbf{Metric}&\textbf{EAS-cluster (Greedy)}\\
    \hline
    $50$& St Dev & 2.4 \%\\
    & Max& 6.86 \%\\
    & Min&-3.87 \%\\
    \hline
    $100$&  St Dev &1.95 \%\\
    & Max& 3.47 \%\\
    & Min&-5.73 \%\\
    \hline
\end{tabular}
    \caption{Statistics on percentage increase in hyper-volume of EAS-cluster with Monte Carlo Simulation relative to Greedy policy for the experiments reported in Table \ref{experimental results 2}}
    \label{add stats 2}
\end{table}

\begin{table}[H]
    \centering
    \begin{tabular}{cc|c}
$\mathbf{n}$&\textbf{Metric}&\textbf{No-EAS}\\
    \hline
    $50$& St Dev &6.57 \%\\
    & Max& 25.7 \%\\
    & Min&-2.71 \%\\
    \hline
    $100$&  St Dev & 4.73 \%\\
    & Max& 16.66 \%\\
    & Min&0.3 \%\\
    \hline
    $200$&  St Dev & 3.53 \%\\
    & Max& 10.99 \%\\
    & Min&-0.87 \%\\
\end{tabular}
    \caption{Statistics on percentage increase in hyper-volume of EAS-cluster relative to NoEAS for the experiments reported in Table \ref{experimental results 3} with more variable travel times.}
    \label{add stats 3}
\end{table}

\end{document}